\documentclass{article}

\usepackage{graphicx}      
\usepackage[authoryear]{natbib}
\setcitestyle{authoryear}
\usepackage{amsmath} 
\usepackage{amssymb}  
\usepackage{url}
\usepackage{authblk}

\usepackage{nicefrac}

\DeclareMathOperator{\nll}{\mathcal{L}}

\newcommand{\R}{\mathbb{R}}
\newcommand{\nin}{{n_u}} 
\newcommand{\ny}{{n_y}} 
\newcommand{\nx}{{n_x}}
\newcommand{\nsamp}{N}
\newcommand{\npar}{{n_\theta}}
\newcommand{\F}{\mathcal{F}} 
\newcommand{\G}{\mathcal{G}} 
\newcommand{\N}{\mathcal{N}} 
\newcommand{\MAP}{{\rm MAP}}
\newcommand{\tvec}[1]{{\mathbf{#1}}}
\newcommand{\mean}[1]{\hat{#1}}

\newcommand{\D}{\mathcal{D}} 
\newtheorem{remark}{Remark}

\author{Marco Forgione} 
\author{Dario Piga} 

\affil{{IDSIA Dalle Molle Institute for Artificial Intelligence USI-SUPSI, Lugano, Switzerland (  
name.surname@supsi.ch).}}

\title{Neural State-Space Models: Empirical Evaluation of Uncertainty Quantification} 

\begin{document}
\maketitle




\begin{abstract}                
Effective quantification of uncertainty is an essential and still missing step towards a greater adoption of deep-learning approaches in 
different applications, including mission-critical ones. In particular, investigations on the predictive uncertainty of deep-learning 
models describing non-linear dynamical systems are very limited to date. This paper is aimed at filling this gap and presents preliminary results on uncertainty quantification for system identification with neural state-space models. We frame the learning problem in a Bayesian probabilistic setting 
and obtain posterior distributions for the neural network's weights and outputs through approximate inference techniques.
Based on the posterior, we construct credible intervals on the outputs and define a \emph{surprise index} which can effectively diagnose usage of the model in a potentially dangerous out-of-distribution regime, where  predictions cannot be trusted.
\end{abstract}

\vskip .5em

\noindent\rule{\textwidth}{1pt}
Please cite this version of the paper:\\ \\
M. Forgione and D. Piga. Neural State-Space Models: Empirical Evaluation of Uncertainty Quantification. In \textit{Proc. of the 22nd IFAC World Congress, Yokohama, Japan}, 2023.
\vskip 1em \noindent
You may use the following bibtex entry:
\begin{verbatim}
@inproceedings{forgione2023empirical,
  title={{Neural State-Space Models: Empirical Evaluation
          of Uncertainty Quantification}},
  author={Forgione, Marco and Piga, Dario},
  booktitle={Proc. of the 22nd IFAC World Congress, Yokohama, Japan},
  year={2023}
}
\end{verbatim}
\noindent\rule{\textwidth}{1pt}

\section{Introduction}
In recent years, the system identification community has shown renewed interest in deep-learning tools and techniques for data-driven modeling of non-linear dynamical systems \citep{ljung2020deep}. 
To cite a few examples, system identification approaches based on 1-D convolutional neural networks are presented in \cite{andersson2019, wu2019deep}.
Training of neural NARX architectures with a regularization term promoting decay of the model's linearized impulse response are introduced in 
\cite{peeters2022narx}. Neural networks architectures and fitting criteria  for continuous-time dynamical model identification are presented in \cite{Mavkov20}. 
Finally, algorithms for efficient training of tailor-made neural state-space models are discussed in \cite{forgione2020model} and \cite{beintema2021nonlinear}. 

A common (and justified) criticism to the above-mentioned deep system identification approaches is the general lack of uncertainty description and analysis. Indeed, the methods presented in those contributions only produce \emph{nominal} point predictions, with no explicit measure of their reliability.
While the models are shown to deliver high performance in the considered benchmarks, results may dramatically deteriorate when they are used in an \emph{out-of-distribution} regime, i.e. on a test set whose characteristics (in terms of input amplitude, frequency, power, etc.) differ significantly from the ones of the training data. Even worse, no mechanism is in place to \emph{detect} this failure mode, and models may quietly produce off-target, possibly dangerous predictions.

In current machine learning research, uncertainty quantification is recognized as paramount to increase reliability and acceptance of black-box models like neural networks, and it is thus seen as a fundamental step towards their adoption in mission-critical
applications~\citep{loquercio2020general}. Different approaches, both deterministic and probabilistic, have been proposed, see \cite{gawlikowski2021survey} for a recent survey. The probabilistic perspective is arguably more general and theoretically appealing. Certain methodologies like ensemble learning~\citep{lakshminarayanan2017simple} and dropout~\citep{srivastava2014dropout}, first introduced in a deterministic settings, are now better understood as approximate inference algorithms in a Bayesian probabilistic framework.

Most of the contributions on uncertainty quantification presented in the deep-learning literature involve static regression problems (typically from the UCI datasets) with feed-forward neural architectures and/or image classification problems (typically from the CIFAR dataset and variants thereof) with convolutional ones, see \cite{maddox2019simple, wilson2020bayesian, izmailov2021bayesian}. To date, little attention has been devoted to sequential learning problems and in particular 
to non-linear dynamical systems modeling.

A notable exception is the recent contribution \citep{zhou2022sparse}, where learning of dynamical systems in neural input/output form is formulated  
in a Bayesian probabilistic framework. Compared to \citep{zhou2022sparse}, our work is focused on neural \emph{state-space} models, which are arguably more suitable for downstream control applications (e.g. for model predictive control) and for analysis with standard system theoretic tools. Furthermore, the main objectives in \citep{zhou2022sparse} are to select the relevant input regressors and to induce sparsity in the network, while our work is focused on uncertainty description and recognition of the out-of-distribution regime. 

We obtain uncertainty bounds by framing the neural state-space identification problem in a Bayesian probabilistic settings and by deriving 
(approximate) posterior distributions for the neural network parameters and for its output predictions.
From a technical perspective, we use the Laplace approximation \citep{bishop2006pattern} to describe the parameter posterior distribution and exploit our recent results 
in \citep{forgione2022adaptation} to speed up the required Hessian matrix computations.
We show that the obtained uncertainty bounds, while not always calibrated, widen significantly when neural state-space models are used in an out-of-distribution regime. 

Based on the obtained uncertainty description, we then introduce a new metric, called \emph{surprise} index that, for a given trained model and a {new} input sequence, detects whether the model is suitable to predict the corresponding output before collecting any new data. 
Thus, the surprise index may be used to assess beforehand whether the predictions generated by a model fed by a specific input
signal can be trusted.

We demonstrate the effectiveness of our methodology on a variation of the Wiener-Hammerstein identification benchmark \citep{schoukens2009wiener} conveniently modified to generate data from different regimes, and release the codes required to reproduce our results in the GitHub repository \url{https://github.com/forgi86/sysid-neural-unc}.

\section{Methodology}
\subsection{Dataset and objective}
We are given a dataset $\D=(\tvec{u}, \tvec{y})$ with $\nsamp$ input samples $u_k \in \mathbb{R}^{\nin}$ and (possibly noisy) output samples $y_k \in \mathbb{R}$, collected from a dynamical data-generating system $\mathcal{S}$. 

Our goal is to estimate from $\D$ a neural state-space model $M$ of the unknown dynamics of $\mathcal{S}$, which for a new input sequence $\tvec{u}^*$, generates a prediction of the corresponding output ${\tvec{y}}^*$, plus an \emph{indicator} of the predictions' reliability. 

Given a suitable probabilistic neural model structure with a prior distribution defined over its parameters, the problem may be tackled through (approximate) statistical inference of the \emph{posterior predictive distribution}  (ppd) $p(\tvec{y}^* | \tvec{u}^*, \D)$, which in turn may be used to generate output predictions with credible intervals. 

In this paper, we follow indeed a probabilistic approach bearing in mind that, due to assumptions and approximations introduced to carry out the inference step efficiently, the obtained ppd and bounds may be somewhat inaccurate. Still, we aim at exploiting probabilistic reasoning and tools to obtain  useful
indicators of model predictions' reliability. These indicators should (at least) be able to detect when the model is operating in an extrapolation
regime, and thus its prediction cannot be fully trusted.

The case of multi-input single-output systems is discussed to simply exposition. However, the results can be extended straightforwardly to multi-input multi-output systems. 

\subsection{Model structure}
We consider the following neural state-space model structure $M$: 
 \begin{subequations}
  \label{eq:ss_model}
 \begin{align}
  x_{k+1} &= \F(x_k, u_k; \theta) \label{eq:ss_model_a} \\
  \mean{y}_{k} &= \G(x_k; \theta) \label{eq:ss_model_b}\\
  y_{k} &=  \mean{y}_{k} + e_k, \qquad e_k \sim \N\left(0, \nicefrac{1}{\beta}\right)  \\
  \theta &\sim \N\left(0, \nicefrac{1}{\tau}\right),
  \end{align}
\end{subequations}
where $\F$ and $\G$ are feed-forward neural networks having compatible dimensions, $x_k \in \mathbb{R}^{\nx}$ is the state at time $k$, and $\theta \in \mathbb{R}^{n_\theta}$ is a vector of parameters to be estimated from data. The measured output $y_{k} \in \R$ is assumed to be corrupted by a zero-mean white Gaussian noise error $e$ with {precision} $\beta$. The prior on the model parameters $\theta$ is also Gaussian, with precision $\tau$.

\section{Probabilistic derivations}
\subsection{Posterior parameter distribution}
The  posterior distribution $p(\theta |  \D)$ of $\theta$ conditioned on the observations $\D$ is given by the Bayes rule:
\begin{equation}
\label{eq:bayes_rule}
p(\theta |  \D) = \frac{p(\theta) p(\D|\theta)}{p(\D)}.
\end{equation}
The functional form of the Gaussian \emph{prior} distribution $p(\theta)$ on the model parameters $\theta$ is:
\begin{equation}
\label{eq:theta_prior}
p(\theta) = \frac{\tau^\npar}{\sqrt{(2\pi)^{\npar} }} \exp\left(\frac{-\tau}{2} \sum_{i=0}^{\npar-1} \theta_i^2\right),
\end{equation}
while the \emph{likelihood} $p(\mathcal{D}|\theta)$ is:
\begin{equation}
\label{eq:likelihood}
p(\mathcal{D}|\theta) = \frac{\beta^\nsamp}{\sqrt{(2\pi)^{\nsamp} }} \exp\left(\frac{-\beta}{2}\sum_{k=0}^{\nsamp-1}{(y_k - \mean{y}_k(\theta))^2}\right),
\end{equation}


\subsection{Posterior predictive distribution}
The posterior predictive distribution $p(\tvec{y}^* | \tvec{u}^*, \D)$ given a new input sequence $\tvec{u}^*$ is:
\begin{equation}
\label{eq:posterior_predictive}
p(\tvec{y}^* | \tvec{u}^*, \D) = \int_{\theta} p(\tvec{y}^* | \tvec{u}^*, \theta) p(\theta | \D) \; d\theta.
\end{equation}
Even when the approximate $p(\theta | \D)$ has a simple structure, exact solution of the integral above is intractable and further approximations/simplifications are required to evaluate the ppd.

\section{Approximate inference}
We present in this section the approximate inference approaches used to obtain the parameter posterior $p(\theta | \D)$ and 
the predictive posterior $p(\tvec{y}^*| \tvec{u}^*, \D)$.
\subsection{Laplace approximation of the parameter posterior}
The parameter posterior $p(\theta | \D)$ is approximated using the Laplace's method \citep{bishop2006pattern} centered around the maximum a posteriori (MAP) point estimate  ${\theta^\MAP}$. All the derivations and required computations are specified in this section.
\subsubsection{MAP point estimate}
To obtain the MAP estimate, we consider the \emph{negative} logarithm of the posterior distribution $\nll(\theta) = - \log p(\theta | \D)$:
\begin{equation}
\label{eq:nll}
 \nll(\theta) = {} \overbrace{\frac{\beta}{2} \sum_{k=0}^{\nsamp-1} (y_k - \mean{y}_k(\theta))^2}^{=E_{\rm lik}(\theta)} +
 \overbrace{\frac{\tau}{2} \sum_{i=0}^{\npar-1} \theta_i^2}^{=E_{\rm prio}(\theta)}
 + \rm{cnst},
\end{equation}
where $\rm{cnst}$ is a term that does not depend on $\theta$.

The MAP estimate ${\theta^\MAP}$ is:
\begin{equation}
  \label{eq:theta_map}
 {\theta^\MAP} = \arg \min_\theta \nll(\theta).
\end{equation}
Computation of $\theta^\MAP$ corresponds to a non-linear (regularized) least-squares problem, which for neural state-space models is usually tackled with
stochastic gradient descent algorithms or variants thereof. 
\subsubsection{Laplace approximation}
The Laplace approximation of the parameter posterior distribution centered around the MAP estimate is defined as:
\begin{equation}
 p(\theta | \D) = \N(\theta^\MAP, P_{\theta^\MAP}),
\end{equation}
where 
$P_{\theta^\MAP}$ is the inverse of the Hessian of 
the {negative }log-likelihood evaluated in $\theta^\MAP$:

\begin{equation}
 P_{\theta^\MAP}^{-1} =  \left .\frac{\partial^2 \nll(\theta)}{\partial \theta^2} \right |_{\theta = {\theta^\MAP} .}
\end{equation}

The Hessian of $E_{\rm prio}(\theta)$ has the simple functional form:
\begin{equation}
 \frac{\partial^2 E_{\rm prio}(\theta)}{\partial \theta^2} 
 = \tau I,
\end{equation}
while the Hessian of $E_{\rm lik}(\theta)$ is:
\begin{equation}
\label{eq:full_hessian}
\frac{\partial^2 E_{\rm lik}(\theta)}{\partial \theta^2}
=  \beta \!\sum_{k=0}^{\nsamp-1} \frac{\partial \mean{y}_k}{\partial \theta} {\frac{\partial \mean{y}_k}{\partial \theta}}^{\top} +
\beta \!\sum_{k=0}^{\nsamp-1} (\mean{y}_k\! -\! y_k) \frac{\partial^2 \mean{y}_k}{\partial \theta^2}.
\end{equation}
According to the  Gauss-Newton (GN) Hessian approximation \citep{wright1999numerical}, the expression above  is dominated by the first term $\beta \!\sum_{k=0}^{\nsamp-1} \frac{\partial \mean{y}_k}{\partial \theta} {\frac{\partial \mean{y}_k}{\partial \theta}}^{\top}$ and the second contribution $\beta \sum_{k=0}^{\nsamp-1} (\mean{y}_k - y_k) \frac{\partial^2 \mean{y}_k}{\partial \theta^2}$ may be neglected.  The GN approximation is accurate, for instance, when $\frac{\partial^2 \mean{y}_k}{\partial \theta^2}$ is small (i.e. model predictions are nearly affine), when $\mean{y}_k - y_k$ is small (which is typically the case for the optimal $\theta$ if the variance of $e_k$ 
is also small) and, more in general, when $\frac{\partial^2 \mean{y}_k}{\partial \theta^2}$ and $\mean{y}_k - y_k$ are uncorrelated (which is also expected
for the optimized value of $\theta$, as the residual $\mean{y}_k - y_k$ is then close to the white measurement noise $e_k$).

Overall, the covariance matrix $P_{\theta^{\MAP}}$ with GN Hessian approximation is:
\begin{equation}
\label{eq:P_post}
P_{\theta^{\rm MAP}}^{-1} \approx \tau I + \beta \!\sum_{k=0}^{\nsamp-1} \frac{\partial \mean{y}_k}{\partial \theta} {\frac{\partial \mean{y}_k}{\partial \theta}}^\top.
\end{equation}

\begin{remark}
The term $\beta \!\sum_{k=0}^{\nsamp-1} \frac{\partial \mean{y}_k}{\partial \theta} {\frac{\partial \mean{y}_k}{\partial \theta}}^\top$ in \eqref{eq:P_post} is also a finite-sample approximation of the \emph{Fisher Information Matrix}, which corresponds in frequentist statistics to the (asymptotic) precision of the maximum likelihood estimator \citep{van2007parameter}. In this sense, the methodologies of this paper are also applicable to the derivation of confidence intervals in a frequentist setting.
\end{remark}

\subsubsection{Computational aspects}A straightforward approach to obtain of the gradients $\frac{\partial \mean{y}_k}{\partial \theta},\; k\!=\!0,\dots,\nsamp\!-\!1$ 
needed in \eqref{eq:full_hessian} is to invoke $\nsamp$ independent back-propagation operations through the neural state-space model's unrolled computational graph at each time step. Overall, this naive approach requires a number of operations $\mathcal{O}(N^2)$. 

The computational cost can actually be lowered to $\mathcal{O}(N)$ using the recursive methodology based on sensitivity equations first introduced by the authors in \cite{forgione2022adaptation} and reported hereafter for completeness.

Let us introduce the \emph{state sensitivities} $s_{k} = \frac{\partial x_{k}}{\partial \theta}\in R^{\nx \times \npar}$.
By taking the derivatives of the left- and right-hand side of~\eqref{eq:ss_model_a} w.r.t. the model parameters $\theta$, we obtain a recursive equation describing the evolution of $s_k$:
\label{eq:sens}
\begin{equation}
    \label{eq:sens_x}
    s_{k+1} = J^{fx}_k s_k + J^{f\theta}_k,
\end{equation}
where $J^{fx}_k \in \R^{\nx \times \nx}$ and $J^{f\theta}_k \in \R^{\nx \times \npar}$ are the Jacobians of  $\mathcal{F}(x_k, u_k; \theta)$ w.r.t. $x_k$ and $\theta$, respectively.

Let us now take the derivative of \eqref{eq:ss_model_b} w.r.t. $\theta$:
\begin{align}
\label{eq:sens_y}
    \frac{\partial \mean{y}_k}{\partial \theta} = J^{gx}_{k} s_k + J^{g\theta}_{k},
\end{align}
where $J^{gx}_{k} \in \R^{\ny \times \nx}$ and $J^{g\theta}_{k} \in \R^{ny \times \npar}$ are the Jacobians of  $\mathcal{G}(x_{k}, u_{k}; \theta)$ w.r.t. $x_{k}$ and $\theta$, respectively,  and $n_y$ is the number of 
outputs ($n_y=1$ in this paper).

The Jacobians $J^{fx}_{k}$ and $J^{f\theta}_{k}$ can be obtained through $\nx$ back-propagation operations through $\F$, thus at cost $\mathcal{O}(\nx \npar)$. Similarly, $J^{gx}_{k}$ and $J^{g\theta}_{k}$ can be obtained through $n_y$ back-propagation operations through $\G$ at cost $\ny \npar$.
Thus, the computational effort required to obtain $\frac{\partial \mean{y}_k}{\partial \theta}$ in ~\eqref{eq:sens_y} (given the previous sensitivity $s_{k-1}$) is $\mathcal{O}((\nx + n_y)\npar)$. Overall, all the derivatives of interest $\frac{\partial \mean{y}_k}{\partial \theta},\; k\!=\!0,\dots,\nsamp\!-\!1$ are then computed at a total cost $\mathcal{O}(\nsamp (\nx + n_y)\npar)$.\\

\subsection{Linearization-based approximation of the ppd}
Our approximation of the posterior predictive distribution is based on a linearization of the neural network model with respect to its parameters about the MAP estimate:
\begin{equation}
 \mean{\tvec{y}}^*(\theta) \approx \mean{\tvec{y}}^*(\theta^{\MAP}) + J^*(\theta - \theta^{\MAP}),
\end{equation}
where $J^{*}$ is the Jacobian of $\mean{\tvec{y}}^*$ with respect to the parameters $\theta$, computed for $\theta=\theta^{\rm MAP}$.
According to the approximation above, we obtain:
\begin{equation}
\label{eq:ppd_approx}
 \tvec{y}^* \sim \N \left(\mean{\tvec{y}}^*(\theta^{\MAP}),\;\; \overbrace{J^* P_{\theta^{\MAP}} {J^*}^\top  + \frac{1}{\beta} I}^{=\Sigma_{\tvec{y}^*}} \right).
\end{equation}
Note that the ppd's covariance matrix is the sum of a term $J^* P_{\theta^{\MAP}} {J^*}^\top$ related to the approximate knowledge of the true system parameters (our epistemic uncertainty), plus a term $\nicefrac{1}{\beta} I$ related to measurement noise (the intrinsic aleatoric uncertainty). 

We are interested in particular into the diagonal entries of $\Sigma_{\tvec{y}^*}$, which correspond to the variance of the output predictions at the different time steps and thus represent their uncertainty. Specifically, we construct and visualize 99.7\% credible intervals centered around the nominal prediction and having width $\pm 3$ times the square root of these diagonal entries.

\subsubsection{Surprise index}
The diagonal entries of $J^* P_{\theta^{\MAP}} {J^*}^\top$ are also of interest and they are related to the variance in the noise-free output predictions. In particular, a relatively large ratio between the $k$-th diagonal entry of the matrix and $\mean{{y}}^*_{k}$ indicates an unreliable prediction at time instant $k$ whose uncertainty is large compared to the predicted value itself. 
For a full sequence $\tvec{u}^*$ of length $\nsamp$, we introduce in this paper an aggregate \emph{surprise} index $s(\tvec{u}^*)$ defined by:
\begin{equation}
\label{eq:surprise}
s(\tvec{u}^*) =  100 \times \frac{\sum_{k=0}^{\nsamp-1} \sqrt{\left (J^* P_{\theta^{\MAP}} {J^*}^\top \right )}_{kk}}{\sum_{k=0}^{\nsamp-1} |\mean{y}_k^{*}(\theta^{\MAP})|}~(\%),
\end{equation}
which measures the relative size of the uncertainty throughout the sequence $\tvec{u}^*$. In \eqref{eq:surprise}, the subscript  $kk$ denotes the diagonal element of a matrix in the $k$-th row and column.   

It is important to note that computation of $s(\tvec{u}^*)$ does not require the actual output $\tvec{y}^*$ and thus it can be carried out even without running an actual experiment on the real system.
Therefore, $s(\tvec{u}^*)$ may be used to assess beforehand whether the model is expected to give reliable predictions when fed with the sequence $\tvec{u}^*$.

\section{Numerical example}
In this section, we test the methodologies presented in the paper on a non-linear system identification problem. 
The developed software is based on the PyTorch deep-learning library and it is available in the GitHub repository \url{https://github.com/forgi86/sysid-neural-unc}.
Computations are performed on a PC equipped with an AMD Ryzen 5 1600x processor, 32 GB of RAM, and an nvidia 1060 GPU.

\begin{figure}
 \centering
 \includegraphics[width=.8\linewidth]{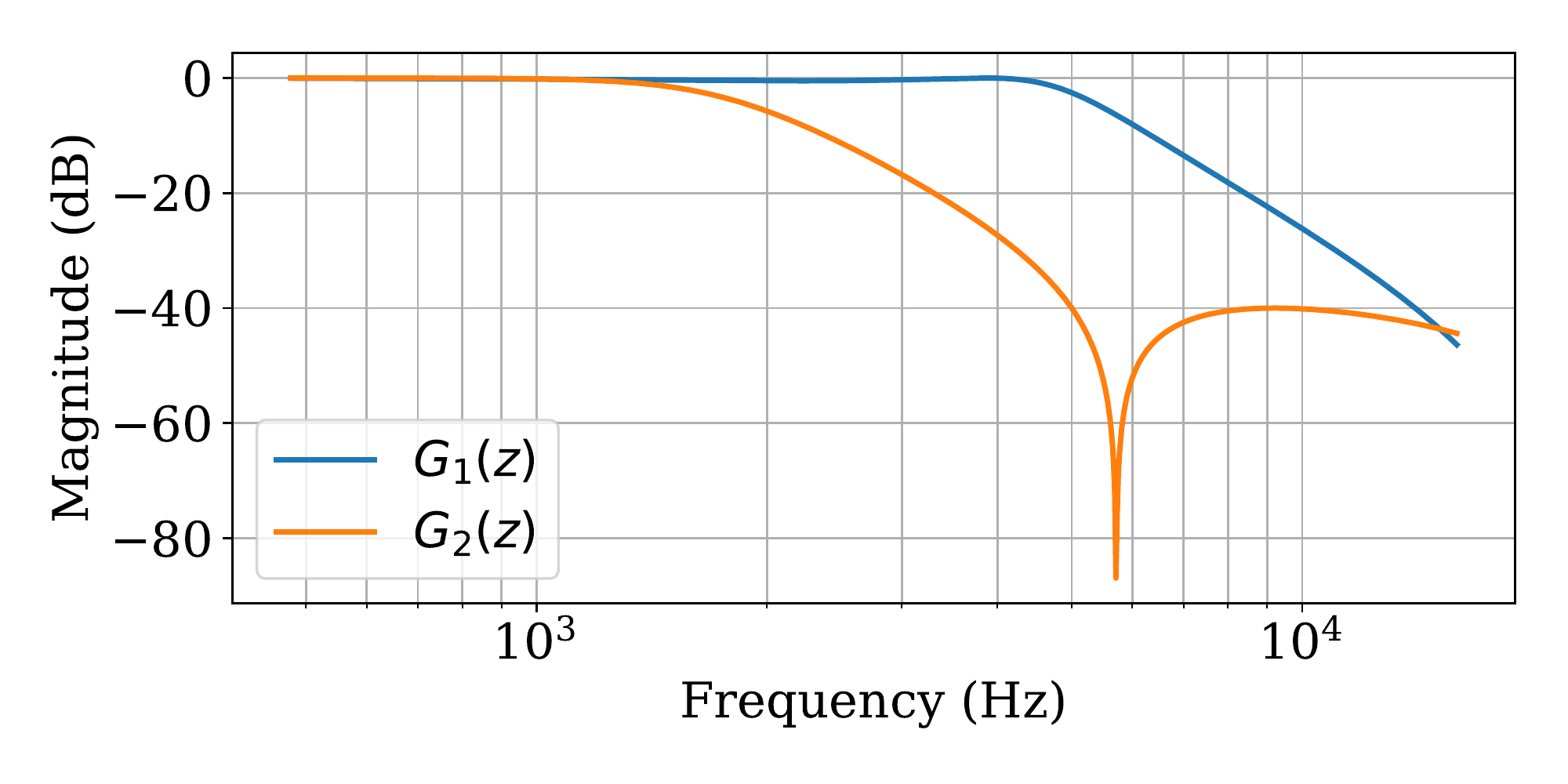}
 \caption{WH system: Bode diagrams of $G_1(z)$ and $G_2(z)$.}
 \label{fig:wh_bode}
\end{figure}

\begin{figure}
 \centering
 \includegraphics[width=.7\linewidth]{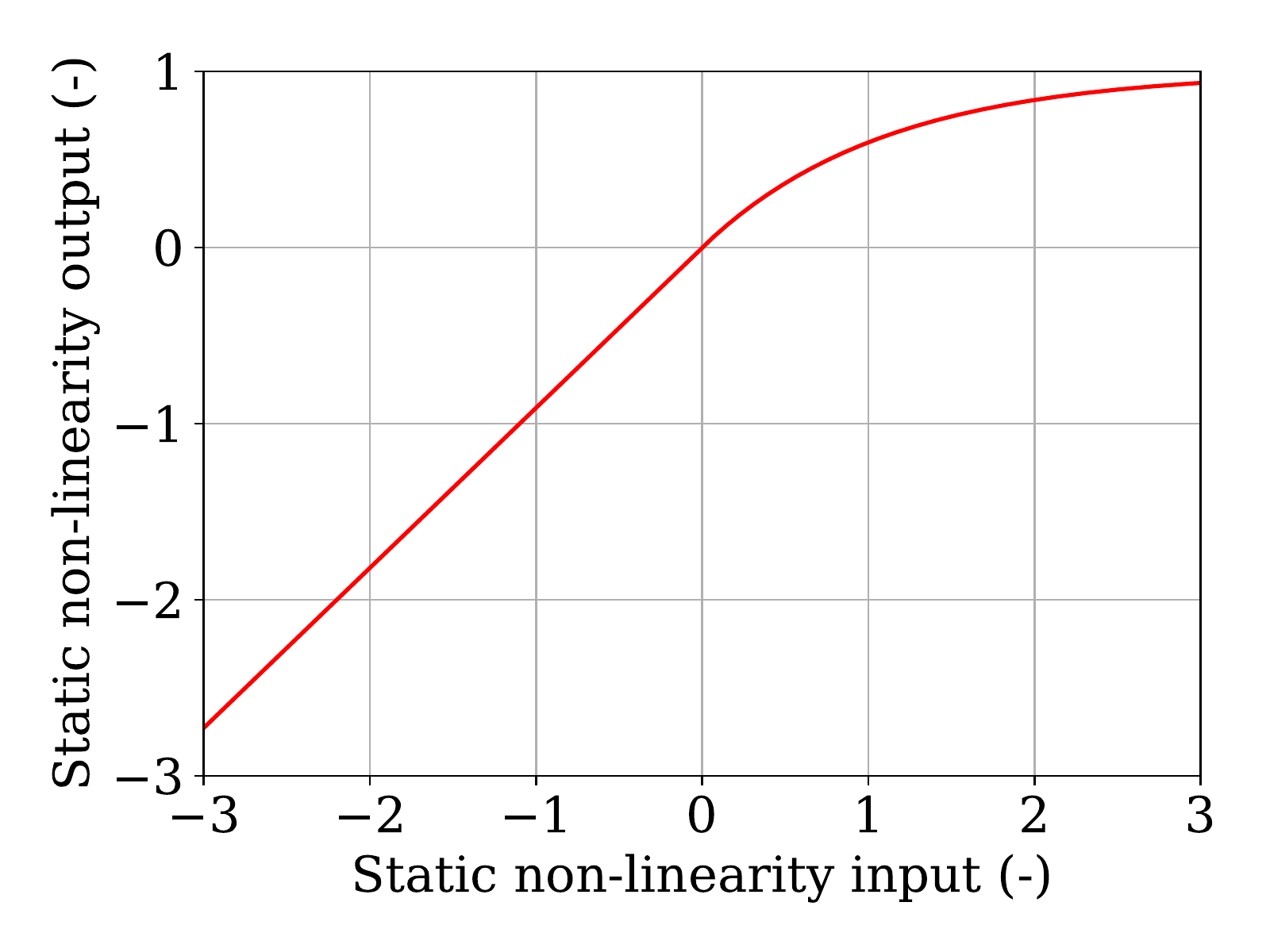}
 \caption{WH system: Static non-linearity $f(\cdot)$.}
 \label{fig:wh_static}
\end{figure}

We consider as true system a discrete-time Wiener-Hammerstein with sampling frequency $f_s=51200$~Hz consisting in the series interconnection of a transfer function $G_1(z)$, a static non-linearity $f(\cdot)$, and a transfer function $G_2(z)$:
\begin{small}
\begin{align*}
 G_1(z) &= \frac{0.010252 + 0.030757z^{-1} + 0.030757z^{-2} + 0.010252z^{-3}}{ 1 -2.151941z^{-1} + 1.744729z^{-2} -0.510767z^{-3}}\\
 G_2(z) &= \frac{0.008706 -0.004596z^{-1} - 0.004596z^{-2} + 0.008706z^{-3}}{1 -2.574867z^{-1} + 2.235716z^{-2} -0.652629z^{-3}}\\
 f(x) &= {\rm elu}\left (-\frac{10}{11}x \right ),
 \end{align*}
\end{small}
where ${\rm elu}(x) = e^{x-1}$ for $x \leq 0$ and $0$ otherwise.
The Bode plots of $G_1$, $G_2$ and the static non-linearity $f(\cdot)$ are also shown in Figure~\ref{fig:wh_bode} and Figure~\ref{fig:wh_static}, respectively.
{
This Wiener-Hammerstein system is closely inspired to the dynamics of the benchmark \citet{schoukens2009wiener} involving a real electronic circuit. In this paper, we prefer this synthetic Wiener-Hammerstein system to the original benchmark in order to be able to generate data from different dynamical regimes with ease.} 
{For analogy with \citet{schoukens2009wiener}, inputs and outputs of our numerical example are assumed to be in Volts (V) units hereafter.}

As for the neural state-space model \eqref{eq:ss_model}, in line with previously published results on the benchmark \citep{beintema2021nonlinear}, $\mathcal{F}$ and $\mathcal{G}$ have a single hidden layer with 15 nodes and $\rm tanh$ static non-linearity, plus a direct linear input/output term. In total, the model has $n_\theta=385$ parameters.

We use a training dataset where the input is a 10000-sample \emph{multisine} signal with flat spectrum 
in the frequency range [0\; 2]~kHz and standard deviation 0.4~V, and the output is corrupted by a white Gaussian noise with standard deviation $\sigma_e=\nicefrac{1}{\beta}=5 \cdot 10^{-3}$~V. Note that the input spectrum does not cover the \emph{transmission zero} of the transfer function $G_2(z)$ located at approximately 5.5~kHz.

To compute the MAP estimate $\theta^\MAP$ efficiently, the negative log-likelihood \eqref{eq:nll} is minimized over batches of sub-sequences extracted from the training data in random order, see \citet{forgione2020model} for details. The batch size and sub-sequence length are both set to 256. 
Neural network parameter optimization is performed over 120 epochs\footnote{An epoch corresponds to the processing of all the contiguous sub-sequences of length 256 in the training dataset in random order.} of the Adam algorithm followed by 4 epochs of L-BFGS, using the standard implementation and default settings of PyTorch. Overall, the optimization procedure takes 873 s.

Once $\theta^\MAP$ is available, the posterior covariance $P_{\theta^\MAP}$ is obtained according to the Laplace approximation \eqref{eq:P_post}. The time required to obtain $P_{\theta^\MAP}$, which is largely dominated by the computation of the gradients $\frac{\partial \mean{y}_k}{\partial \theta}$, is  44 s using the recursive gradient computation method in \cite{forgione2022adaptation}, while it increases to 465 s with the naive implementation.

For model testing, we consider four  scenarios where the input signal $\tvec{u}^*$ is:
 1) a multisine with standard deviation 0.4~V and bandwidth [0\; 2]~kHz (same as training input);
 2) a multisine with standard deviation 0.4~V and bandwidth [1\; 2]~kHz;
 3) a multisine with standard deviation 0.8~V and bandwidth [0\; 2]~kHz;
 4) a multisine with standard deviation 0.4~V and bandwidth [0\; 10]~kHz.
 For each test set, we compute the nominal prediction $\mean{\tvec{y}}^*$ by simulating the state-space model \eqref{eq:ss_model} with $\theta=\theta^\MAP$ and  $e=0$, and the approximate ppd according to \eqref{eq:ppd_approx}. The approximate ppd is then used to obtain 99.7\% credible intervals (having width $\pm 3$ times the square root of the diagonal entries of the approximate ppd's covariance matrix $\Sigma_{\tvec{y}^*}$) and the surprise index $s(\tvec{u}^*)$ according to \eqref{eq:surprise}.

We evaluate the performance of the nominal predictions in terms of the FIT index:
\begin{equation}
\label{eq:fit_index}
\mathrm{FIT} = 100 \times \left(1- \frac{\sqrt{\sum_{k=0}^{\nsamp-1} \left({y}^*_k -  \mean{y}^*_k\right)^2} }  
{\sqrt{\sum_{k=0}^{\nsamp-1} \left({y}^*_k -  {\overline{y}^*}\right)^2}}\right) (\%),
\end{equation}
where ${\overline{y}^*}$ is the sample mean of the sequence $\tvec{y}^*$. 

To evaluate the goodness of the credible intervals, we report their empirical \emph{coverage}, namely the percentage 
of time steps where the actual output $\tvec{y}^*$ lies inside the intervals. A value close to $99.7\%$ indicates \emph{well-calibrated} intervals.

Note that Signals 3 and 4 drive the system in a dynamical ranges unseen during training and thus force the model to operate in an extrapolation regime. For these signals, we expect the FIT index to decrease and the uncertainty intervals to get wider. Wider uncertainty bounds  result in a larger surprise index $s(\tvec{u}^*)$, which in turn should allow us to detect the FIT decrease without knowledge of the actual output $\tvec{y}^*$.

The FIT index, surprise index, and coverage of the four test signals are reported in Table~\ref{tab:wh_results}. We observe that FIT and surprise indexes are indeed negatively correlated, as expected. 

\begin{figure}
 \centering
 \includegraphics[width=.99\linewidth]{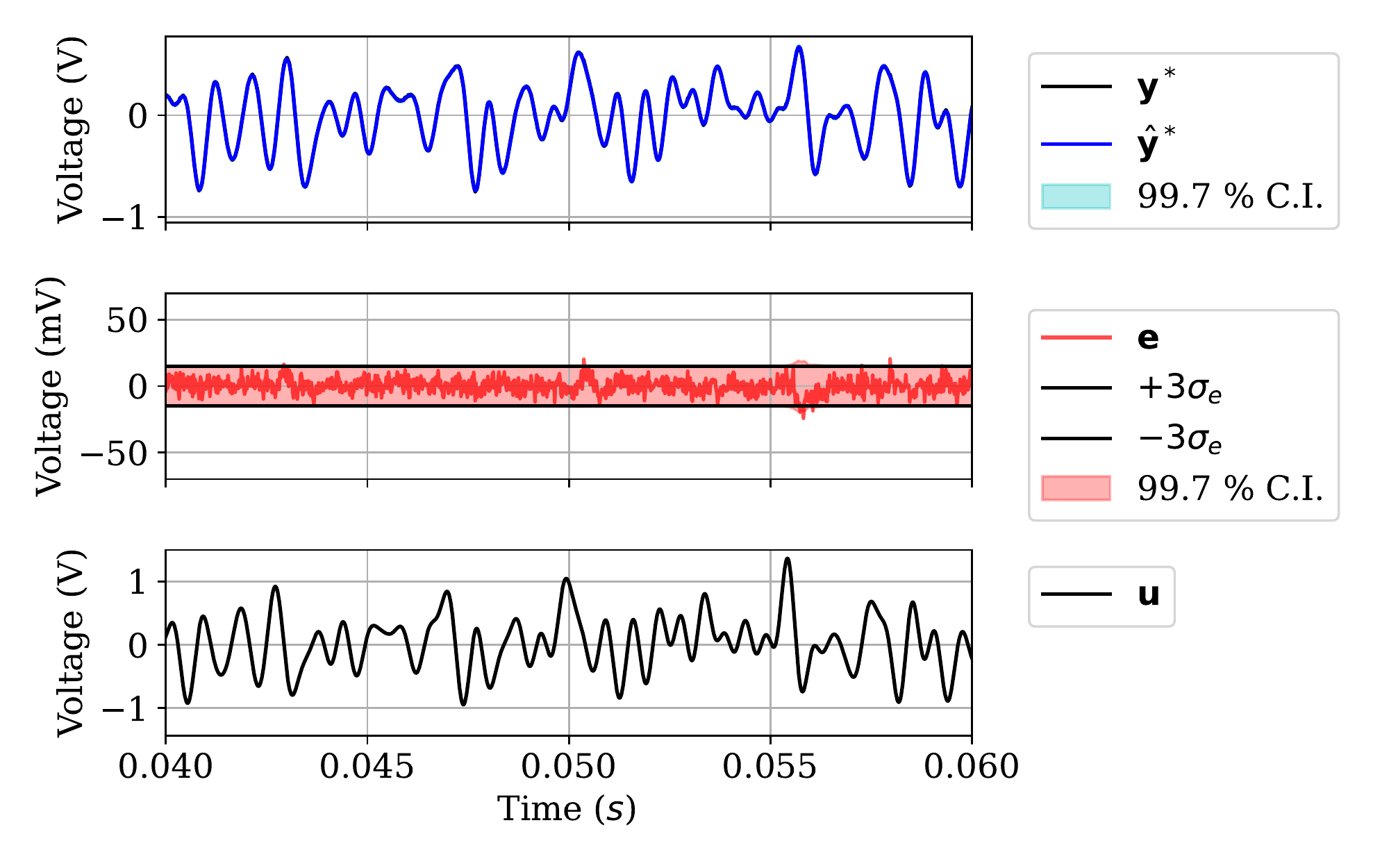}
 \caption{WH system: results on multisine signal 1.}
 \label{fig:wh_multisine_1}
\end{figure}

\begin{figure}
 \centering
 \includegraphics[width=.99\linewidth]{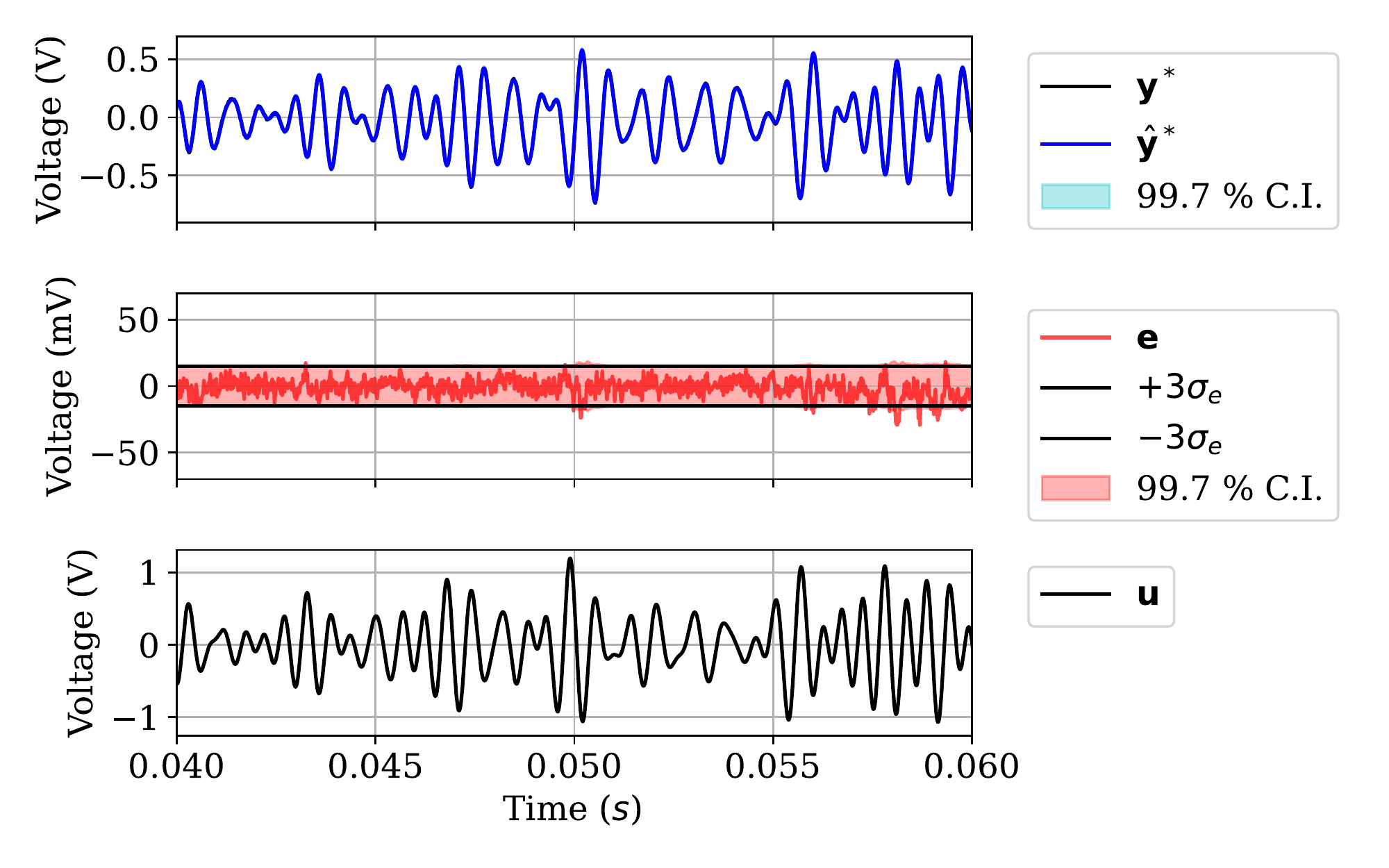}
 \caption{WH system: results on multisine signal 2.}
 \label{fig:wh_multisine_2}
\end{figure}

\begin{figure}
 \centering
 \includegraphics[width=.99\linewidth]{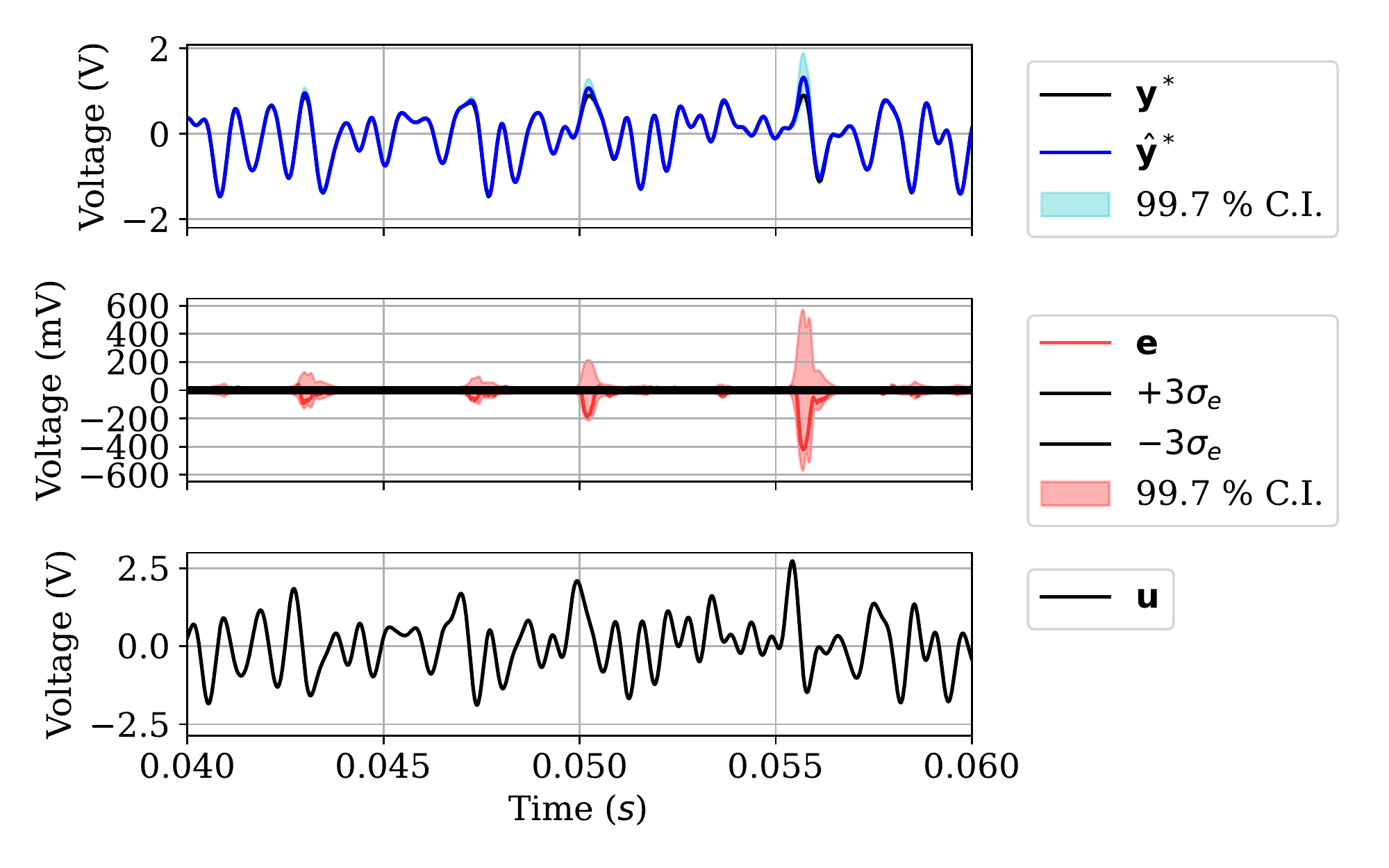}
 \caption{WH system: results on multisine signal 3.}
 \label{fig:wh_multisine_3}
\end{figure}

\begin{figure}
 \centering
 \includegraphics[width=.99\linewidth]{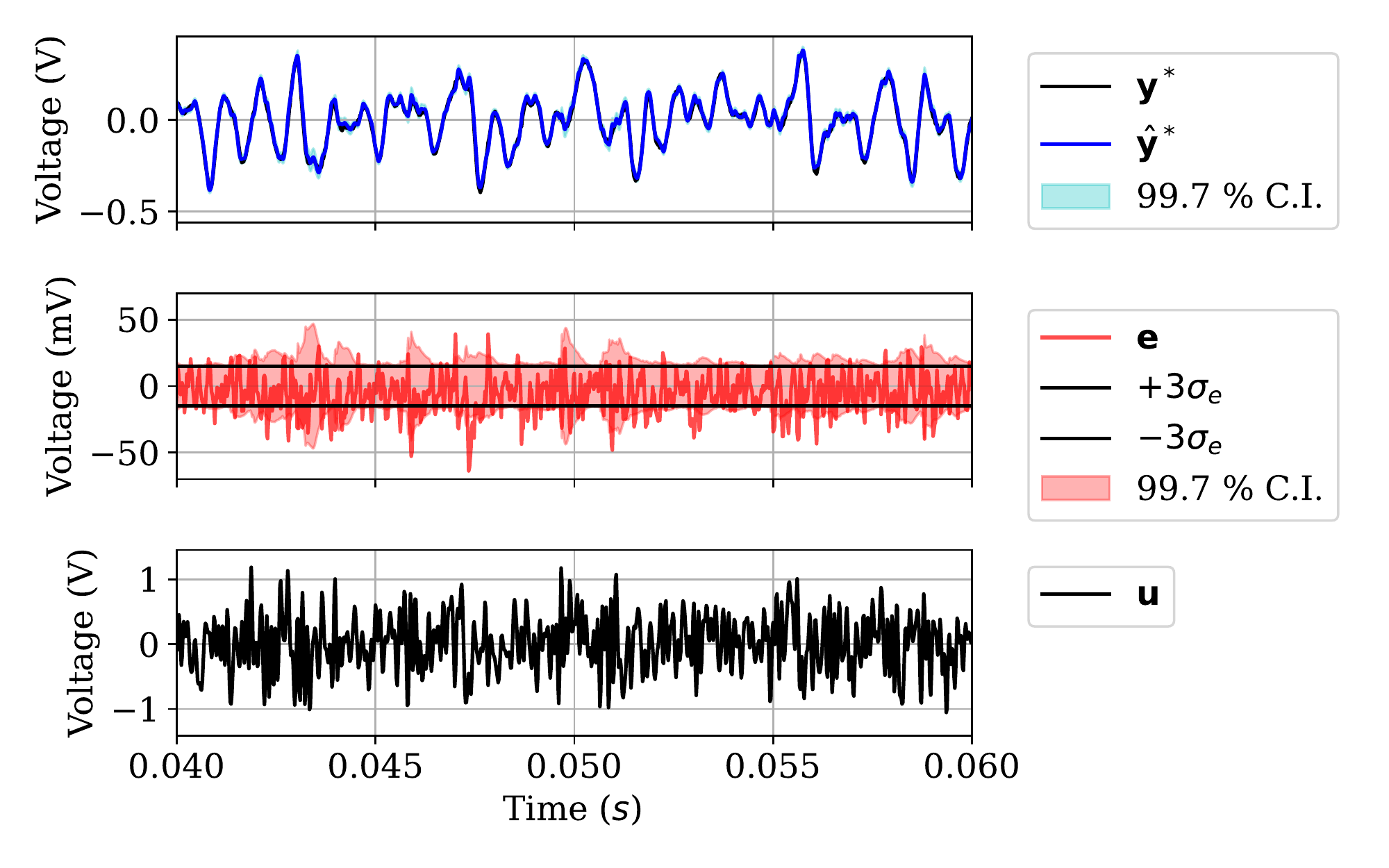}
 \caption{WH system: results on multisine signal 4.}
 \label{fig:wh_multisine_4}
\end{figure}

In Figures \ref{fig:wh_multisine_1}, \ref{fig:wh_multisine_2}, \ref{fig:wh_multisine_3}, \ref{fig:wh_multisine_4}, we show relevant time traces for the four test signals. In the top panel, we show the actual output $\tvec{y}^*$ (black line) together with the posterior mean $\mean{\tvec{y}}^*$ (blue line) and 99.7\% credible intervals (shaded blue area). In the middle panel, we show the error signal $\tvec{e}= \tvec{y}^*-\mean{\tvec{y}}^*$ (red line), together with 99.7\% credible intervals (shaded red area). We also show  $\pm 3 \sigma_e$ horizontal bands, corresponding to the aleatoric component of the output uncertainty  (black lines). Finally, the bottom panel represents the input signal. 

For Signals 1 and 2, the prediction quality is very high (black and blue line overlapping in the the top panel). Uncertainty bounds are not visible (as too narrow) in the top panel and can only be appreciated in the (magnified) middle one. In the middle panel, we also note that uncertainty bounds are largely dominated by the aleatoric component (red shaded area mostly comprised within the $\pm 3 \sigma_e$ bands). Furthermore, these bounds are well calibrated (the red line lies within the bounds for most of the time steps), as also indicated by the coverage indexes which are close to the target value $99.7$~\%.

For Signal 3, the prediction quality decreases significantly in the time instant when the input/output samples are large (a condition not seen during training). The uncertainty bounds also expand in these regions and, remarkably, appear to be still rather well calibrated (i.e. the error $\tvec{e}$ lies indeed within the uncertainty region in most of the time steps, with a coverage index of 96.1~\%). Furthermore, the surprise index of 2.10~\% is significantly larger than the one computed over Signals 1 and 2 (0.33~\% and 0.43~\%, respectively).
Thus, the out-of-distribution regime is effectively detectable using the proposed methodology.

For Signal 4, uncertainty bounds are also enlarged, even though they are clearly not well calibrated. Indeed, the error signal $\tvec{e}$ is 
too often out of the red shaded area, thus the latter is not a well-calibrated 99.7\% credible interval, as also indicated by the low coverage of 80.6~\%). Nonetheless, the high surprise index $s(\tvec{u}^*)=4.03$  alerts the used that the model is working in an extrapolation regime and consequently its performance will be low.

\begin{table}
    \centering
    \begin{tabular}{|l||l|l|l|}
    \hline
    Signal &  FIT (\%) & coverage (\%) & surprise (\%)\\
    \hline
    multisine 1 &98.1 & 99.2 & 0.33\\
    multisine 2 &97.7 & 98.6 & 0.43\\
    multisine 3 &93.9 & 96.1 & 2.10\\
    multisine 4 &87.8 & 80.6 & 4.03  \\
    \hline
    \end{tabular}
    \caption{FIT index, uncertainty intervals coverage, and surprise index on the test datasets.}
    \label{tab:wh_results}
\end{table}
\section{Conclusion}
We have presented a viable approach for uncertainty quantification with neural state-space models. Based on the obtained uncertainty description, we have defined a surprise index that indicates whether the model predictions generated from a given input are expected to be reliable, i.e. close to the response of the true system.

This preliminary work may be extended in different directions. First, other inference approximation techniques may be adopted to obtain a richer and more accurate characterization of the uncertainty. In this sense, efficient sampling techniques such as Hamiltonian Monte Carlo may be considered to overcome the limiting assumptions (e.g. uni-modality) of the currently used Laplace approximation. A challenge in this sense is to devise scalable algorithms applicable to large neural-network models.

Furthermore, tools like the surprise index may be used 
to {choose} informative input signals to be used for model training/refinement. 
This could pave the way for experiment design and active learning in the context of system identification with neural state-space models.

Finally, to foster further research in uncertainty quantification and out-of-distribution recognition, specific benchmarks and performance metrics should be devised and shared with the system identification community.

\section*{Acknowledgement}
This work was partially supported by the European H2020-CS2 project ADMITTED, Grant agreement no. GA832003.

\bibliography{references}                           

\begin{thebibliography}{20}
\providecommand{\natexlab}[1]{#1}
\providecommand{\url}[1]{\texttt{#1}}
\providecommand{\urlprefix}{URL }
\expandafter\ifx\csname urlstyle\endcsname\relax
  \providecommand{\doi}[1]{doi:\discretionary{}{}{}#1}\else
  \providecommand{\doi}{doi:\discretionary{}{}{}\begingroup
  \urlstyle{rm}\Url}\fi

\bibitem[{Andersson et~al.(2019)Andersson, Ribeiro, Tiels, Wahlström, and
  Schön}]{andersson2019}
Andersson, C., Ribeiro, A.H., Tiels, K., Wahlström, N., and Schön, T.B.
  (2019).
\newblock {D}eep {C}onvolutional {N}etworks in {S}ystem {I}dentification.
\newblock In \emph{2019 IEEE 58th Conference on Decision and Control (CDC)},
  3670--3676.
\newblock \doi{10.1109/CDC40024.2019.9030219}.

\bibitem[{Beintema et~al.(2021)Beintema, T{\'o}th, and
  Schoukens}]{beintema2021nonlinear}
Beintema, G., T{\'o}th, R., and Schoukens, M. (2021).
\newblock Nonlinear state-space identification using deep encoder networks.
\newblock In \emph{Learning for Dynamics and Control}, 241--250. PMLR.

\bibitem[{Bishop and Nasrabadi(2006)}]{bishop2006pattern}
Bishop, C.M. and Nasrabadi, N.M. (2006).
\newblock \emph{Pattern recognition and machine learning}, volume~4.
\newblock Springer.

\bibitem[{Forgione et~al.(2022)Forgione, Muni, Piga, and
  Gallieri}]{forgione2022adaptation}
Forgione, M., Muni, A., Piga, D., and Gallieri, M. (2022).
\newblock On the adaptation of recurrent neural networks for system
  identification.
\newblock \emph{arXiv preprint arXiv:2201.08660}.

\bibitem[{Forgione and Piga(2020)}]{forgione2020model}
Forgione, M. and Piga, D. (2020).
\newblock Model structures and fitting criteria for system identification with
  neural networks.
\newblock In \emph{2020 IEEE 14th International Conference on Application of
  Information and Communication Technologies (AICT)}, 1--6.
\newblock \doi{10.1109/AICT50176.2020.9368834}.

\bibitem[{Gawlikowski et~al.(2021)Gawlikowski, Tassi, Ali, Lee, Humt, Feng,
  Kruspe, Triebel, Jung, Roscher et~al.}]{gawlikowski2021survey}
Gawlikowski, J., Tassi, C.R.N., Ali, M., Lee, J., Humt, M., Feng, J., Kruspe,
  A., Triebel, R., Jung, P., Roscher, R., et~al. (2021).
\newblock A {S}urvey of {U}ncertainty in {D}eep {N}eural {N}etworks.
\newblock \emph{arXiv preprint arXiv:2107.03342}.

\bibitem[{Izmailov et~al.(2021)Izmailov, Vikram, Hoffman, and
  Wilson}]{izmailov2021bayesian}
Izmailov, P., Vikram, S., Hoffman, M.D., and Wilson, A.G.G. (2021).
\newblock {W}hat are {B}ayesian {N}eural {N}etwork {P}osteriors {R}eally
  {L}ike?
\newblock In \emph{International conference on machine learning}, 4629--4640.
  PMLR.

\bibitem[{Lakshminarayanan et~al.(2017)Lakshminarayanan, Pritzel, and
  Blundell}]{lakshminarayanan2017simple}
Lakshminarayanan, B., Pritzel, A., and Blundell, C. (2017).
\newblock {S}imple and {S}calable {P}redictive {U}ncertainty {E}stimation using
  {D}eep {E}nsembles.
\newblock \emph{Advances in {N}eural {I}nformation {P}rocessing {S}ystems}, 30.

\bibitem[{Ljung et~al.(2020)Ljung, Andersson, Tiels, and
  Sch{\"o}n}]{ljung2020deep}
Ljung, L., Andersson, C., Tiels, K., and Sch{\"o}n, T.B. (2020).
\newblock {D}eep {L}earning and {S}ystem {I}dentification.
\newblock \emph{IFAC-PapersOnLine}, 53(2), 1175--1181.

\bibitem[{Loquercio et~al.(2020)Loquercio, Segu, and
  Scaramuzza}]{loquercio2020general}
Loquercio, A., Segu, M., and Scaramuzza, D. (2020).
\newblock A {G}eneral {F}ramework for {U}ncertainty {E}stimation in {D}eep
  {L}earning.
\newblock \emph{IEEE Robotics and Automation Letters}, 5(2), 3153--3160.

\bibitem[{Maddox et~al.(2019)Maddox, Izmailov, Garipov, Vetrov, and
  Wilson}]{maddox2019simple}
Maddox, W.J., Izmailov, P., Garipov, T., Vetrov, D.P., and Wilson, A.G. (2019).
\newblock A {S}imple {B}aseline for {B}ayesian {U}ncertainty in {D}eep
  {L}earning.
\newblock \emph{Advances in Neural Information Processing Systems}, 32.

\bibitem[{Mavkov et~al.(2020)Mavkov, Forgione, and Piga}]{Mavkov20}
Mavkov, B., Forgione, M., and Piga, D. (2020).
\newblock Integrated neural networks for nonlinear continuous-time system
  identification.
\newblock \emph{IEEE Control Systems Letters}, 4(4), 851--856.

\bibitem[{Peeters et~al.(2022)Peeters, Beintema, Forgione, and
  Schoukens}]{peeters2022narx}
Peeters, L., Beintema, G.I., Forgione, M., and Schoukens, M. (2022).
\newblock {NARX} identification using {D}erivative-{B}ased {R}egularized
  {N}eural {N}etworks.
\newblock \emph{arXiv preprint arXiv:2204.05892}.

\bibitem[{Schoukens et~al.(2009)Schoukens, Suykens, and
  Ljung}]{schoukens2009wiener}
Schoukens, J., Suykens, J., and Ljung, L. (2009).
\newblock {W}iener-{H}ammerstein benchmark.
\newblock In \emph{Proc. of the 15th {IFAC} symposium on System Identification
  (SYSID 2009)}.

\bibitem[{Srivastava et~al.(2014)Srivastava, Hinton, Krizhevsky, Sutskever, and
  Salakhutdinov}]{srivastava2014dropout}
Srivastava, N., Hinton, G., Krizhevsky, A., Sutskever, I., and Salakhutdinov,
  R. (2014).
\newblock Dropout: {A} {S}imple {W}ay to {P}revent {N}eural {N}etworks from
  {O}verfitting.
\newblock \emph{The {J}ournal of {M}achine {L}earning {R}esearch}, 15(1),
  1929--1958.

\bibitem[{Van~den Bos(2007)}]{van2007parameter}
Van~den Bos, A. (2007).
\newblock \emph{Parameter estimation for scientists and engineers}.
\newblock John Wiley \& Sons.

\bibitem[{Wilson and Izmailov(2020)}]{wilson2020bayesian}
Wilson, A.G. and Izmailov, P. (2020).
\newblock {B}ayesian {D}eep {L}earning and a {P}robabilistic {P}erspective of
  {G}eneralization.
\newblock \emph{Advances in neural information processing systems}, 33,
  4697--4708.

\bibitem[{Wright et~al.(1999)Wright, Nocedal et~al.}]{wright1999numerical}
Wright, S., Nocedal, J., et~al. (1999).
\newblock Numerical optimization.
\newblock \emph{Springer Science}, 35(67-68), 7.

\bibitem[{Wu and Jahanshahi(2019)}]{wu2019deep}
Wu, R.T. and Jahanshahi, M.R. (2019).
\newblock Deep {C}onvolutional {N}eural {N}etwork for {S}tructural {D}ynamic
  {R}esponse {E}stimation and {S}ystem {I}dentification.
\newblock \emph{Journal of Engineering Mechanics}, 145(1), 04018125.

\bibitem[{Zhou et~al.(2022)Zhou, Ibrahim, Zheng, and Pan}]{zhou2022sparse}
Zhou, H., Ibrahim, C., Zheng, W.X., and Pan, W. (2022).
\newblock Sparse {B}ayesian {D}eep {L}earning for {D}ynamic {S}ystem
  {I}dentification.
\newblock \emph{Automatica}, 144, 110489.

\end{thebibliography}

\end{document}